# Component-Based Machine Learning for Indoor Flow and Temperature Fields Prediction: Latent Feature Aggregation and Flow Interaction

Shaofan Wang [a], Nils Thuerey [b], Philipp Geyer [a]

[a] Sustainable building system, Leibniz Universität Hannover, Herrenhäuser Str. 8, 30419 Hannover, Germany

[b] School of Computation, Information and Technology, Technical University Munich, Boltzmannstraße 3, 85748 Garching, Germany

**Abstract.** Accurate and efficient prediction of indoor airflow and temperature distributions is essential for building energy optimization and occupant comfort control. However, traditional CFD simulations are computationally intensive, limiting their integration into real-time or design-iterative workflows. This study proposes a component-based machine learning (CBML) surrogate modeling approach to replace conventional CFD simulation for fast prediction of indoor velocity and temperature fields. The model consists of three neural networks: a convolutional autoencoder with residual connections (CAER) to extract and compress flow features, a multilayer perceptron (MLP) to map inlet velocities to latent representations, and a convolutional neural network (CNN) as an aggregator to combine single-inlet features into dual-inlet scenarios. A two-dimensional room with varying left and right air inlet velocities is used as a benchmark case, with CFD simulations providing training and testing data. Results show that the CBML model accurately and fast predicts two-component aggregated velocity and temperature fields across both training and testing datasets. For 95% of the flow field, the maximum absolute error is below 0.08 m/s in velocity prediction and 0.4 °C in temperature prediction. Compared to monolithic surrogate models, the modular CBML structure enhances model transparency and flexibility while maintaining accuracy. Additionally, t-SNE visualization of the latent space confirms the model's ability to differentiate and aggregate inlet-specific features in a physically consistent manner. This approach offers a promising solution for real-time indoor environmental prediction and provides novel opportunities for the integration of data-driven prediction in iterative processes of building design and control.

## 1. Introduction

The attention on indoor hygiene and comfort has explosively increased since the COVID-19 pandemic resulting in an urgent demand for the rapid prediction of indoor environment [1]. As airborne transmission of pathogens became a critical concern, the ability to monitor and anticipate the concentration and movement of indoor air over time and space has proven to be essential for ensuring a hygienic and safe indoor environment [2][3]. Besides, to enhance indoor human comfort, especially under varying occupancy and external weather conditions, well-considered design and precise control of HVAC systems is necessary. The foundation of such control lies in the accurate prediction of the indoor thermal and humidity environment [4]. Moreover, in the pursuit of sustainable development, the realization of passive houses and net-zero carbon buildings has become a key objective. Achieving these energy-efficient building standards requires reduction of ventilation and its energy losses connected to a thorough understanding of the spatiotemporal distribution of indoor air velocity, temperature, and contaminants [5].

Thus, whether for hygiene, energy efficiency, or comfort, prediction of indoor environmental variables has become a fundamental requirement in modern building design and operation. There are many methods are employed to predict the indoor environment. Temperature and flow form key mechanisms of indoor environment prediction. Traditional engineering analytical models of energy simulation used for demand prediction, such as EnergyPlus [6], typically yield only an average temperature value, failing to capture the detailed spatial variations throughout the space as they are based on a systems approach representing an indoor space's air volume as a node [7]. Such simplifications are inadequate for many practical applications where localized thermal conditions are critical or where energy-efficient natural ventilation is desired [8]. To overcome this limitation, Computational Fluid Dynamics (CFD) simulations have become an essential tool over the past decades, offering high-resolution insights into the spatial and temporal characteristics of indoor thermal environments [9]. In temperature prediction, CFD has been widely used to simulate buoyancy-driven ventilation and convective heat transfer [10]. For example, a researcher conducted full-scale CFD simulations of a three-story office building with natural ventilation and validated the results against experimental data, finding temperature deviations of less than 0.3°C, demonstrating the high accuracy of traditional CFD in real buildings [11]. In addition, traditional CFD techniques have been extensively used to model the spread of airborne contaminants such as $CO_2$ and VOCs under various ventilation conditions. A recent study utilized CFD to simulate inter-unit pollutant transfer in dormitory buildings, revealing the influence of wind direction on pollutant transport pathways and providing insights for indoor air quality management [12]. Many buildings today still suffer from being "sick buildings," characterized by excessive energy consumption, poor indoor comfort, and inaccurate temperature and humidity control [13]; in case of local pollutant sources, air flow prediction is very helpful in design. This raises the question: what are the underlying reasons for the 'sick building'? Due to the high computational cost and time-consuming nature of CFD simulations, CFD simulation and analysis are typically conducted only after the architectural design or construction phase is complete, and even then, only a limited number of simulations can be performed. As a result, CFD is due to high modelling effort and computational load normally unable to provide real-time feedback to designers regarding whether their design meets indoor environmental requirements or whether the HVAC control strategies can achieve precise indoor temperature adjustment. The idea of computational steering relying on this real-time feedback requires parallel/high performance computing [14,15]. Ideally, integrating CFD simulations into the early design process would allow designers to understand the environmental implications and outcomes of each design decision. However, the substantial time and computational resources required for traditional CFD poses challenges to incorporate into the iterative design phase. Although techniques such as computational steering have made real-time interaction more feasible in some contexts, it is still impractical to expect design teams to wait hours or even days for CFD simulation results before proceeding to the next design step. Therefore, a significant challenge is to reduce the computational effort of CFD simulations. Addressing this issue is critical to enabling the integration of CFD into the iterative design process where rapid feedback is essential [16]. Furthermore, the CFD model is based on Navier-Stokes equations and numerical schemes, which are require a high degree of expertise in fluid mechanics and their numerical solutions. As architecture designers and building engineers usually do not have this professional knowledge, the application of CFD is often error-prone due to incorrect parameter settings. [17].

To overcome these barriers, simplified CFD models are proposed and investigated. The coarse-grid CFD [18,19] has the potential to deliver results at real-time speeds, offering valuable insights for decision-making and control, its reliability remains a subject of concern due to the uncertain magnitude and distribution of numerical errors introduced by the reduced spatial resolution. Moreover, fast fluid dynamics (FFD) [20,21] is applied to indoor environment prediction. For a typical indoor prediction case, this method reduces the computing time from 464.8 h with CFD to 7.6 h; however, it is limited by its relatively low spatial resolution and simplified numerical schemes, which compromise the accuracy of flow predictions, particularly in regions with strong gradients or complex turbulence structures. Moreover, recent studies have demonstrated the effectiveness of GPU-accelerated lattice Boltzmann methods (LBM) for indoor airflow and temperature distributions, showing good agreement with traditional Large Eddy Simulation approaches while reducing computational time [22]. However, even with a certain reduction in accuracy, the computational cost remains unacceptably high. Therefore, a design-integrated, low-effort, intuitive, minimal specialized knowledge, and rapid prediction surrogate model is necessary.

With respect to this goal, machine learning and reduced order modelling methods are introduced into this field to form a surrogate for CFD simulation and achieve fast prediction of the indoor environment [23]. Some linear projection methods, such as proper orthogonal decomposition (POD) and dynamic mode decomposition (DMD) are developed to reduce the calculation effort [24]. POD method is applied to predict ventilation performance in the high-speed train compartment [25]. However, the inherent linear nature of these low-rank approximations limits their ability to accurately reconstruct airflow distributions with high levels of nonlinearity [26]. Afterward, various non-linear methods are also employed, such as deep neural network (NN) to predict and reconstruct the flow fields. A neural network is developed for reproducing CFD simulations of non-isothermal indoor airflow fields with different pre-processing methods learning rules between different inlet velocities and airflow patterns [27]. A variational autoencoder (VAE)-based approach is applied for reconstructing and rapidly predicting 3D heat and mass transfer, enhancing modeling efficiency for complex physical fields. The results demonstrate that the method achieves high accuracy while significantly reducing computational cost, showing strong potential for engineering applications [28]. To reduce CFD computational costs, a surrogate model using sensor inputs has been developed for predicting indoor flow fields, but its performance strongly depends on sensor placement and shows limited generalization to rooms with different layouts, such as varying inlet/outlet or heater positions [29]. A hybrid physics-informed and data-driven model with transfer learning was proposed for indoor temperature prediction [30], achieving high efficiency, but its generalization is limited when applied to rooms with different thermal boundary configurations. We applied autoencoder and residual network to reduce the dimensionality and reconstruct the indoor air velocity field with different heights of inlet positions [31]. The results show the potential of this method for real-time prediction with acceptable accuracy.

However, resulting data-driven models created through such methods are valid only for the flow settings and the spatial configuration defined by the training data; parameters can be varied but not topological settings such as number of inlets, outlets, windows and etc. In practical architecture design phase, designers and engineers always add or remove different components to the building and room to develop the design, then evaluate the indoor environment performance. In the process, this type of model, which we call monolithic, is unable to give a real-time prediction as feedback for designers, as its prediction just bases on fixed component type and number. This means that the monolithic surrogate model has limited reusability, flexibility, and generalization. Due to this reason, we are developing the approach of component-based machine learning (CBML) for indoor environment prediction. The concept of CBML is already employed in building energy prediction [32], but there are no precedents of applying CBML to spatial indoor velocity and temperature prediction to increase the flexibility and generalization ability of the surrogate model.

In this paper, a CBML method for spatial flow and temperature field prediction based on CFD simulations is proposed. Key for this development is the aggregation of two interacting flow streams predicted as components by data-driven methods; whereas data-driven prediction has been successfully performed [31], the aggregation is an important breakthrough for CBML in fluid dynamics, as it allows for combining predicted flow fields. The benchmark employed in this paper is a 2D rectangular room, where the variable is inlet number and velocity. There are three conditions, only left inlet, only right inlet and both left and right inlet. Two inlets are regarded as two components, which are predicted by data-driven methods separately. Respectively, then two of separate predictions are combined by the aggregator to achieve the prediction a room with two components. The output of this three-component surrogate model is the indoor velocity field.

The rest of this paper is organized as follows. First, we introduce the engineering constraints and research context, emphasizing the challenges in indoor airflow prediction and the need for surrogate modelling. Next, we describe the methodologies, including the CFD simulation setup and surrogate modelling framework, followed by a detailed discussion of CAER, MLP, and CNN in capturing spatial patterns of velocity and temperature distributions. Then, we present the results and discussion, evaluating model accuracy, computational efficiency, and generalization under different airflow conditions. Finally, we summarize the key findings and outline future research directions, including potential improvements in efficient aggregation algorithms.

## 2. Methodology

In this study, a CBML approach is developed to enable real-time prediction of indoor velocity and temperature fields under multi-inlet conditions using a multi-component aggregation surrogate model. The overall framework of the CBML method with its dataflows illustrated in Figure 1, comprises three core neural network modules: (1) a CAER serving as a feature extractor to reduce the order and extract the latent feature from CFD-simulated flow fields; (2) an MLP acting as a predictor to map inlet boundary condition (i.e., inlet velocities) of individual components (e.g., left or right air supply) to their corresponding latent representation; and (3) a CNN as feature aggregator, which fuse the latent features of each single-component and captures the interaction between them. The framework begins with CFD simulations applied to generate the high-fidelity dataset for model training and evaluation. Then CAER is first trained to encode and decode flow fields, extracting representative latent features for each individual component. Subsequently, the MLP is used to map the inlet velocity of each component (left inlet and right inlet) into a latent feature space. Afterwards, for multi-inlet conditions, the latent features of the left and right inlets are aggregated using the CNN-based aggregator, and the resulting fused representation is decoded to reconstruct the full-field velocity and temperature distributions. In addition, the latent features of different component type, left inlet (blue), right inlet (orange), and dual inlets (green) will be clustered using the t-SNE method to reveal the underlying patterns. Specifically, section 2.1 presents the CFD model set-up and case parameter. Then, the hyperparameter and structure of surrogate model are illustrated in section 2.2.

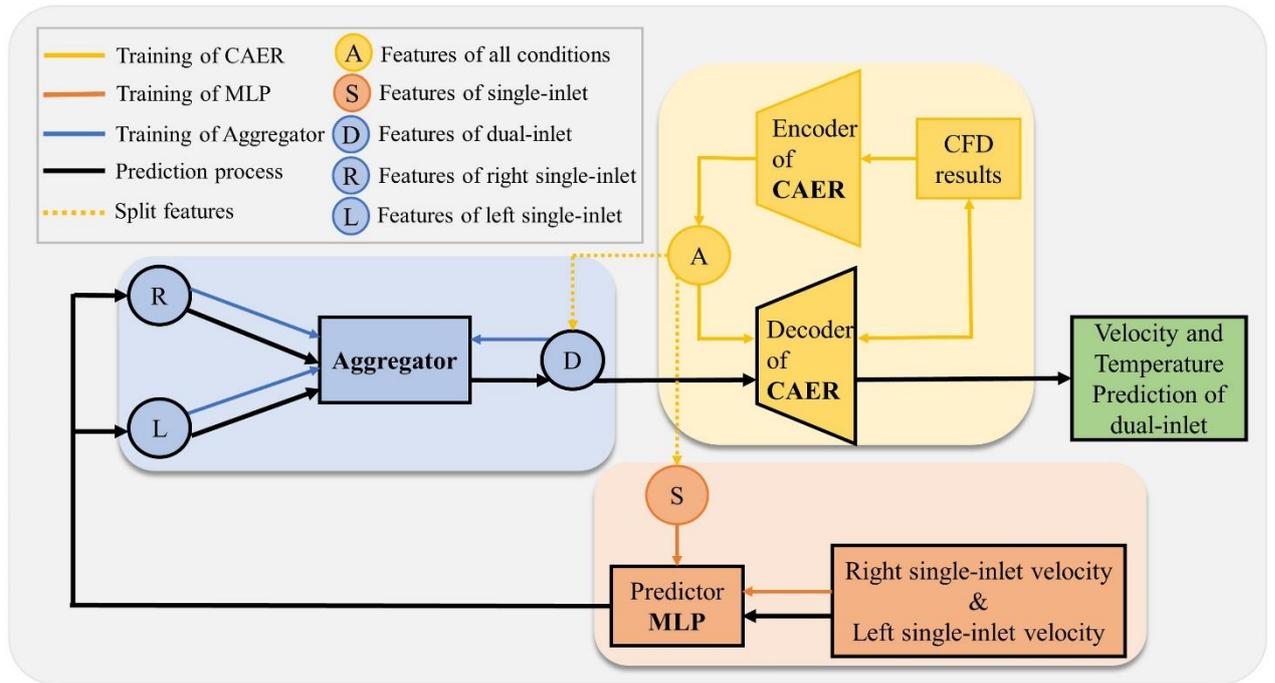

Figure 1: Overview of the dataflows in the CBML Framework

## 2.1 CFD simulation and dataset

In order to generate the dataset for training and evaluate the proposed surrogate model, CFD simulations are utilized to obtain the high-fidelity indoor velocity and temperature distributions under various inlet positions and velocities. In this regard, a two-dimensional (2D) rectangular room served as a benchmark shown in Figure 2 has dimensions of 1.5 meters in the x direction and 1.0 meter in the y direction. Within this room, two windows are present, each measuring 0.9 meters, located on the left and right walls at 35 °C. The thermal boundary conditions of the ceiling and floor are subjected to a uniform heat flux of -200 W meaning heat flow out of room. The temperature of the supply air is set to 10 °C to reflect cool air delivery from the HVAC system. The setting of temperature and heat flux aims to simulate the cooling scenario typically encountered in summer environments.

To systematically analyse both individual and combined ventilation effects, three types of room configurations were applied: (a) left single-inlet, (b) right single-inlet, and (c) dual-inlet (left and right inlet used simultaneously). The first two configurations were used to isolate and extract the latent features corresponding to each inlet independently, while the dual-inlet case was designed to examine the combined effect and potential nonlinear interactions between the two ventilation sources. This setup allows for a component-based modelling approach, where complex flow patterns under multi-inlet conditions are constructed by aggregating the learned features of individual components.

The CFD simulations were performed using OpenFOAM, an open-source finite volume-based solver, under the assumption of steady-state incompressible flow. The Reynolds-Averaged Navier–Stokes (RANS) equations were solved in conjunction with the standard k-ε turbulence model, incorporating buoyancy effects via the Boussinesq approximation. This model offers a practical balance between prediction accuracy and computational efficiency for indoor airflow simulations [33]. The The computational platform is AMD Ryzen Threadripper PRO 3975WX 32-Core processors. The indoor domain is discretized using structured mesh consisting of 15000 cells, ensuring adequate resolution for capturing the flow field details. Besides, to enhance accuracy of turbulence modelling in the near-wall region, wall functions are implemented for all solid boundaries, such as walls, windows, floor and ceiling. These wall functions effectively handle the boundary conditions for turbulent kinetic energy and Reynolds stress in turbulence models, facilitating a more precise calculation of turbulence parameters in near-wall flow regions and ultimately improving simulation accuracy.

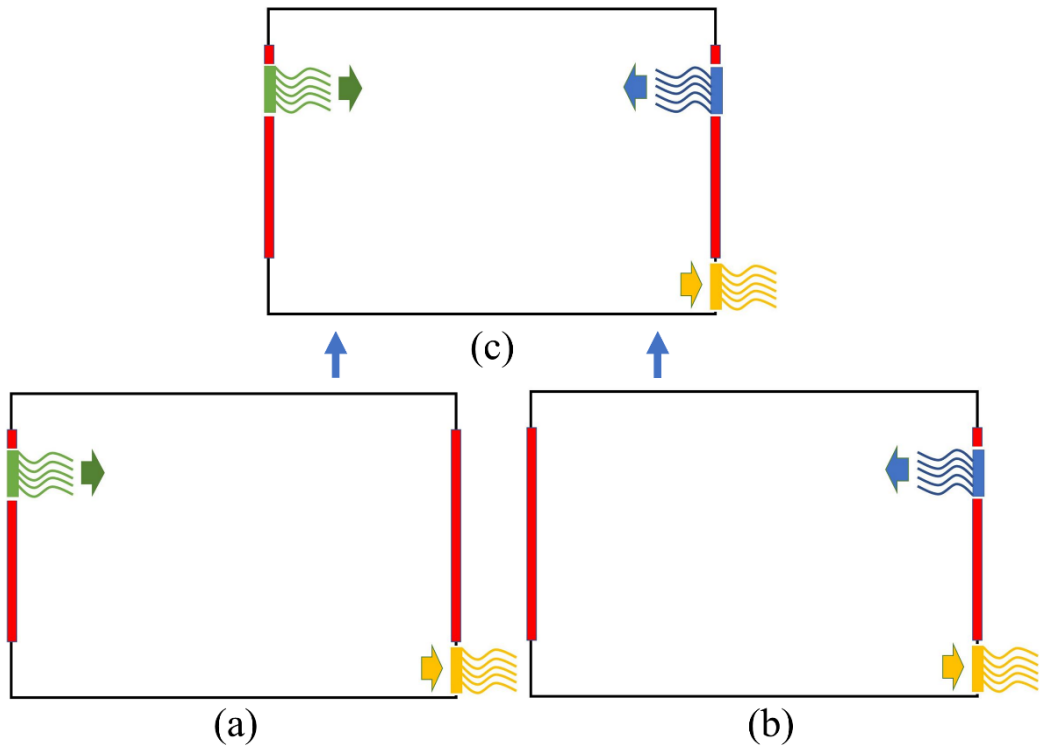

Figure 2. Benchmark case. **(a):** Left single-inlet only; **(b):** Right single-inlet only; **(c):** Dual-inlet.

The training data is specially designed to support the surrogate model in learning and prediction the indoor airflow fields under dual air supply conditions by aggregating airflow fields from single left and single right air supply scenarios. The dataset consists of five different air supply velocity levels for both left and right inlets, covering a realistic range of supply speeds typically observed in building ventilation systems. By combining these velocity levels in a full-factorial manner, a total of 25 dual-inlet cases were generated. For each case, the 2D velocity and temperature fields are extracted and stored as the model's target output. The selection of inlet velocities, presented in Table 1 and Table 2, was carried out manually to ensure sufficient diversity in flow patterns while keeping the dataset size manageable. Although larger datasets can generally improve surrogate model accuracy, they also increase data preparation and training time. Therefore, a trade-off was made to retain computational efficiency and focus the study on evaluating the feasibility and performance of the proposed learning framework with a limited but representative dataset.

Table 1. Training dataset (Unit: m/s).

| Only left inlet | | | | |
|---|---|---|---|---|
| 0.05 | 0.25 | 0.50 | 0.70 | 1.00 |
| Only right inlet | | | | |
| 0.05 | 0.25 | 0.50 | 0.70 | 1.00 |
| Dual inlet (L: left, R: right) | | | | |
| L: 0.05 R: 0.05 | L: 0.05 R: 0.25 | L: 0.05 R: 0.50 | L: 0.05 R: 0.70 | L: 0.05 R: 1.00 |
| L: 0.25 R: 0.05 | L: 0.25 R: 0.25 | L: 0.25 R: 0.50 | L: 0.25 R: 0.70 | L: 0.25 R: 1.00 |
| L: 0.50 R: 0.05 | L: 0.50 R: 0.25 | L: 0.50 R: 0.50 | L: 0.50 R: 0.70 | L: 0.50 R: 1.00 |
| L: 0.70 R: 0.05 | L: 0.70 R: 0.25 | L: 0.70 R: 0.50 | L: 0.70 R: 0.70 | L: 0.70 R: 1.00 |
| L: 1.00 R: 0.05 | L: 1.00 R: 0.25 | L: 1.00 R: 0.50 | L: 1.00 R: 0.70 | L: 1.00 R: 1.00 |

Table 2. Testing dataset (Unit: m/s).

| Dual inlet (L: left, R: right) | | | | | |
|---|---|---|---|---|---|
| L: 0.1 R: 0.9 | L: 0.2 R: 0.8 | L: 0.4 R: 0.6 | L: 0.6 R: 0.4 | L: 0.8 R: 0.2 | L: 0.9 R: 0.1 |

To ensure the reliability of the CFD model before the surrogate model training process, validation should be performed against experimental data. Since the primary focus of this study is to development of CBML surrogate model rather than CFD simulation itself. Therefore, experimental study is not conducted in this study, and the validation in this paper is applied against experimental data reported in [34]. To maintain consistency, the CFD model setup of validation case is same as experimental case in those references. The comparison of u-velocity profiles at mid-plane position, along with the computational domain, is presented in Figure 3. The average relative error was below 5%, indicating strong agreement.

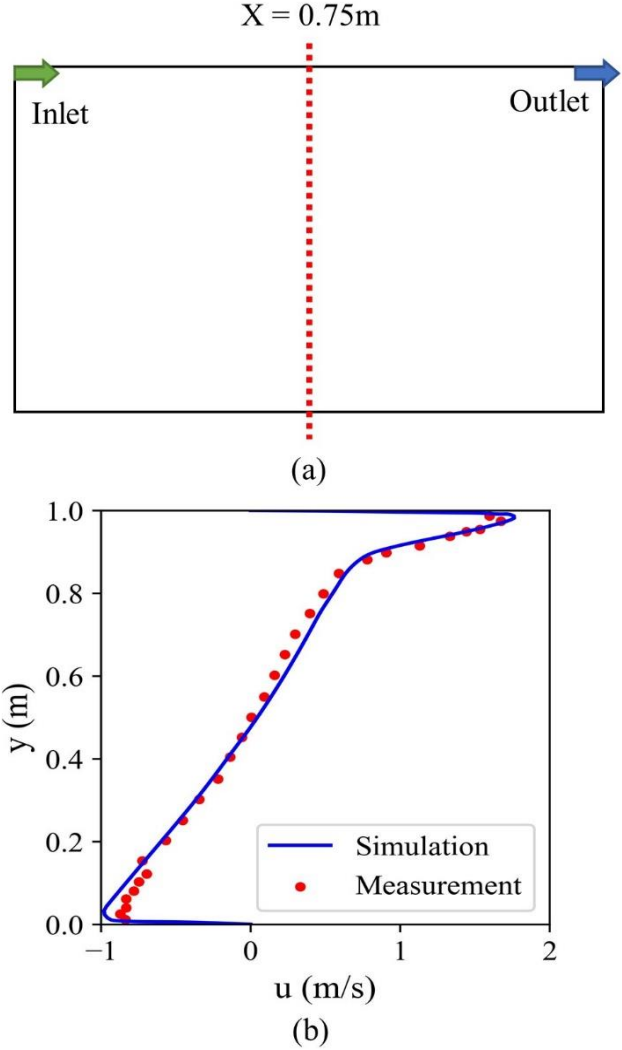

(a)

(b)

Figure 3. (a) Schematic of the validation case geometry, showing a rectangular domain with an inlet on the left and an outlet on the right. The vertical dashed line at x = 0.75 m indicates the location of velocity profile extraction. (b) Comparison of u-velocity profiles at the mid-plane (x = 0.75 m) between the CFD simulation results and experimental measurements from [34].

## 2.2 Surrogate Model

To approximate the computationally expensive CFD simulations under varying ventilation conditions, a component-based surrogate modelling framework is developed. This surrogate model aims to rapidly predict indoor velocity and temperature fields by learning from high-fidelity CFD data and leveraging modular neural network architectures. As briefly outlined in the methodology overview (Section 2), the surrogate model consists of three key components: a CAER to extract compact latent features from individual component flows; an MLP that maps inlet boundary conditions (i.e., inlet air velocities) to corresponding latent features; and a CNN that aggregates latent features from different components and captures the interaction between them in multi-inlet cases. Unlike conventional surrogate models trained end-to-end on full-field flow data, the proposed architecture decomposes the modelling process into interpretable sub-tasks aligned with the physical flow structure. Specifically, the CAER learns spatial patterns of airflow and temperature fields associated with single-inlet conditions, while the MLP generalizes this knowledge by establishing a mapping from boundary input to latent representation of single-inlet condition. The CNN aggregator serves as a feature fusion module that enables prediction of composite flow fields when multiple inlets are simultaneously active. This modular structure offers significant advantages in terms of data efficiency, generalizability, and physical interpretability. In the following paragraphs, the architectures and hyperparameters of each module are described in detail.

### 2.2.1 Convolutional Autoencoder with Residual Network

To extract compact and high-level representations of CFD-simulated indoor flow fields, a CAER is constructed shown in Figure 4. The CAER is designed to reduce the dimensionality of the input velocity and temperature fields while preserving their essential spatial and thermal features. This latent representation serves as the foundation for both boundary condition mapping and multi-component feature aggregation in the surrogate modelling framework. The input to the CAER is trained by cases of left and right single-inlet and dual-inlet. The input is a two-channel field consisting of normalized velocity magnitude and temperature, with a spatial resolution of $100 \times 150$. The encoder comprises five convolutional stages, each consisting of a 2D convolutional layer with kernel size $3 \times 3$, followed by a ReLU activation, a $3 \times 3$ max-pooling layer (stride = 2, padding = 1), and a residual block. The number of channels increases progressively across the layers: from 2 in the input to 16, 32, and finally 64 in the deeper layers. After each pooling layer, the spatial resolution is halved, reducing the feature map to a final latent tensor of size $64 \times 4 \times 5$, making the extracted feature 23.4 times smaller than the original data size. This compressed representation retains the dominant spatial structures and thermal gradients of the original flow field. Each convolutional operation in the encoder is defined mathematically as:

$$y = \sigma\,(W * x + b)$$

where x is the input feature map, $W$ and $b$ are the learnable kernel weights and biases, $*$ denotes the convolution operator, and $\sigma$ is the ReLU activation function.

To improve the network's ability to extract and reconstruct complex flow features, residual blocks are integrated throughout the encoder and decoder of the CAER. Each residual block consists of four convolutional layers with $3 \times 3$ kernels and ReLU activations, structured in two consecutive pairs. The block takes an input tensor $x$, processes it through the convolutional layers to form a transformed output $F(x)$, and then adds the input via a skip connection:

$$y = \sigma\,(F(x) + x)$$

where $F(x)$ represents the output of the convolutional sequence. This identity mapping allows the model to preserve low-level features while learning deeper representations. It also facilitates more stable training by mitigating vanishing gradients and degradation issues commonly found in deeper networks. The use of residual connections is particularly important in this application, as it supports the accurate reconstruction of spatial structures such as recirculation zones and temperature stratification layers within the flow field.

The decoder is divided into two branches, each dedicated to reconstructing either the velocity or temperature field. The shared encoder and separate decoder architecture offer several advantages in predicting indoor airflow fields. First, it extracts common features from both velocity and temperature fields, enhancing feature efficiency. Second, it reduces model complexity and prevents overfitting by lowering the number of parameters. Third, it captures the inherent coupling between velocity and temperature, improving physical consistency. Last but not least, independent decoders allow detailed reconstruction, optimizing specific characteristics of each field. The decoding process mirrors the encoder, consisting of bilinear up-sampling layers followed by transposed convolutions and residual blocks. The spatial resolution is progressively restored to $100 \times 150$, matching the input resolution. The final layer of each decoder applies a Sigmoid activation function:

$$\hat{x} = \frac{1}{1 + e^{-z}}$$

where z is the output of the final transposed convolution layer. This activation ensures that predicted values remain within the normalized range [0, 1], consistent with the pre-processed input data. In this way, the model avoids unbounded outputs and facilitates stable training.

The CAER is trained using supervised learning with paired CFD simulation data as ground truth. The loss function is defined as the sum of the mean squared errors (MSE) between predicted and true velocity and temperature fields:

$$\mathcal{L} = MSE(\hat{V}, V) + MSE(\hat{T}, T)$$

where $\hat{V}, \hat{T}$ denote the predicted fields and V, T represent the ground truth. Training is performed using the Adam optimizer with a learning rate of $1 \times 10^{-4}$ and weight decay of $1 \times 10^{-7}$.

Once trained, the encoder of the CAER is retained and used as a latent feature extractor for single-inlet flow cases. Meanwhile, the decoder is utilized to reconstruct the velocity and temperature fields from the aggregated latent features, enabling the surrogate model to produce full-field predictions under both single- and multi-inlet conditions. The decoder thus serves as the final stage in the surrogate model, translating compressed latent representations into physically interpretable outputs. These features are used in the downstream modules of the surrogate model for boundary-to-latent mapping and feature aggregation.

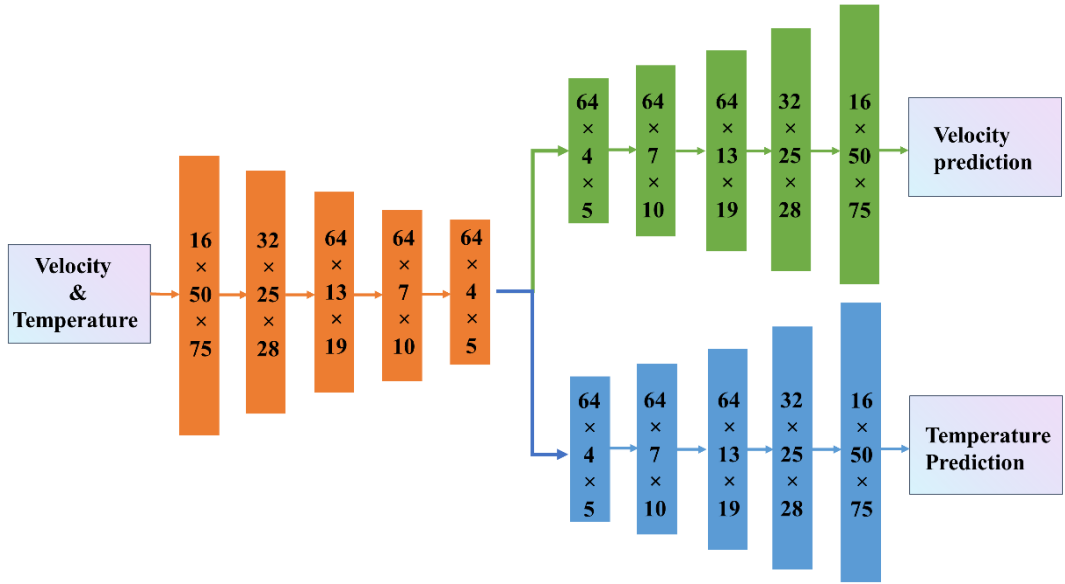

Figure 4. CAER structure with separated decoder.

### 2.2.2 Predictive MLP

To establish a direct and efficient mapping from the inlet boundary conditions to the corresponding latent flow representations for each inlet component, a MLP is employed. This module serves as a predictor, enabling the prediction of low-dimensional latent features solely based on inlet air velocities, without requiring full-field CFD simulation data. The input to the MLP is a two-dimensional vector with format (position, velocity) representing the normalized supply velocities at the left and right inlets. The output is a flattened latent vector of dimension 1280, corresponding to the reshaped feature tensor (64×4×5) extracted by the encoder of the CAER network. The MLP comprises six fully connected layers, with progressively increasing and then stabilizing neuron counts: 128, 512, 1026, 2000, 2000, and finally 1280. Each hidden layer is followed by a ReLU activation function, which introduces nonlinearity and facilitates effective gradient propagation across the deep architecture. This design allows the MLP to approximate complex nonlinear relationships between inlet conditions and the associated latent representations of the indoor flow field. Once trained, the MLP can directly predict the latent vector corresponding to any arbitrary inlet condition within the sampled range, enabling rapid prediction and flow field synthesis when combined with the decoder. The MLP is trained in a supervised fashion using paired datasets $(I_i, O_i)$, where $I_i$ denotes the inlet condition vector and $O_i$ represents the ground-truth latent vector derived from the CAER encoder. The training objective is to minimize the MSE between the predicted and true latent features.

### 2.2.3 CNN-based Aggregator

To model the interaction between multiple ventilation components (e.g., left and right air supply inlets), a CNN-based aggregator is designed to fuse their latent features into a unified representation suitable for full-field flow reconstruction. This module takes as input the concatenated latent vectors of two components and transforms them into a single, physically meaningful latent feature via a series of convolutional layers and residual connections. Specifically, the input to the aggregator is a tensor of size 128×4×5, formed by channel-wise concatenation of two latent vectors of size 64×4×5, each extracted from a single-inlet condition. The aggregator comprises three convolutional stages. The first stage expands the feature dimension to 256 channels using a 3×3 convolution, followed by a ReLU activation and a residual block with two pairs of convolutional layers. This allows the network to model higher-order interactions between the input components. The second stage reduces the number of channels to 128, again followed by a residual block. Finally, the output is compressed to a 64×4×5 latent tensor, matching the shape required for decoding. The aggregator is trained in a supervised manner using ground-truth latent features extracted from CFD simulations under dual-inlet conditions. The loss function is defined as the MSE between the predicted fused latent feature and the true latent representation obtained by encoding the dual-inlet CFD flow field:

$$\mathcal{L}_{agg} = MSE(\hat{z}_{aggregated}, z_{CFD})$$

Where, $\hat{z}_{aggregated}$ is the aggregator output, and $z_{CFD}$ is the encoded latent feature of the dual-inlet CFD flow field. The model is trained using the Adam optimizer with a learning rate $1 \times 10^{-4}$ of and a weight decay of $1 \times 10^{-7}$, for a total of 6000 epochs.

## 3. Results and discussion

In this section, the indoor airflow fields and temperature distributions of the CFD simulation and CBML surrogate model are presented, compared and evaluated on training dataset and test dataset. Performance is assessed through qualitative comparisons with ground truth CFD fields, as well as quantitative error metrics. Specifically, section 3.1 presents the reconstruction performance on the training dataset, while Section 3.2 discusses the model's generalization capability under unseen inlet conditions. Each section is further divided into two parts, corresponding to velocity field (Sections 3.1.1 and 3.2.1) and temperature field (Sections 3.1.2 and 3.2.2) evaluation, respectively.

### 3.1 Prediction on training dataset

The reconstruction performance of the surrogate model for indoor airflow fields and temperature distribution is shown and evaluated on the training dataset consisting of five representative dual-inlet configurations selected from all training database. In these cases, the left inlet velocity is fixed at 0.50 m/s, while the right inlet velocity varies from 0.05 m/s to 1.00 m/s, allowing for an assessment of the model's learning capability under a range of flow interactions through both qualitative visualization and quantitative error metrics.

### 3.1.1 Velocity Field

Figure 5 presents a comparison of CFD results (the first column), surrogate model predictions (the second column) and spatial absolute error maps (the third column) for different inlet conditions in the selected five representatives. In general, the surrogate model effectively captures the global flow structures, including the primary velocity distributions, recirculation zones, and jet interactions with an error not exceeding 0.04 m/s substantially. The predicted velocity fields closely resemble the CFD simulations, demonstrating that the model has successfully learned the dominant flow patterns from the training data. Minor deviations between the predicted and CFD results are evident in certain regions, particularly in areas with strong velocity gradients and shear layers. The observed velocity distributions indicate that increasing the right inlet velocity from 0.05 m/s to 1.0 m/s while left inlet velocity is fixed at 0.50 m/s increases the interaction between the two inlet streams leading to more pronounced velocity gradients and vortex formations. For cases with small right inlet velocities (e.g., L = 0.5 m/s, R = 0.05 m/s and L = 0.5 m/s, R = 0.25 m/s), the surrogate model exhibits noticeable deviations near the jet boundary and recirculation zones. The absolute error maps indicate that the largest discrepancies occur in regions where velocity transitions sharply, suggesting that the model may struggle to capture fine-scale turbulent structures. These errors could be attributed to the inherent complexity of turbulent flows, as well as possible limitations in the model's ability to learn highly nonlinear interactions present in the flow field. In contrast, the surrogate model performs significantly better for balanced inlet conditions (e.g., L = 0.5 m/s, R = 0.5 m/s), where the velocity distributions in both CFD and predicted results show strong alignment. The error maps reveal that difference in these cases is more evenly distributed and generally lower in magnitude, indicating that the model works well on scenarios where momentum contributions from both inlets are relatively equal.

To further compare the reconstruction accuracy across all inlet conditions, Figure 6 presents violin plots of the absolute error distribution, providing a statistical overview of the model's performance over the entire training dataset including both the density and spread of errors across the spatial domain. The plots reveal that the error distributions are consistently narrow and concentrated around low values. Median errors in all cases are below 0.01 m/s, and the 95% of the errors lies within 0.015 m/s. Longer tails in the violin plots, particularly for the most asymmetric cases, reflect localized outliers in regions with strong shear or recirculation, which are inherently more difficult to reconstruct. Importantly, these outliers constitute only a small fraction of the total domain. The strong agreement with CFD results both in visual structure and statistical plot highlights the model's ability to learn complex flow dynamics within the training set.

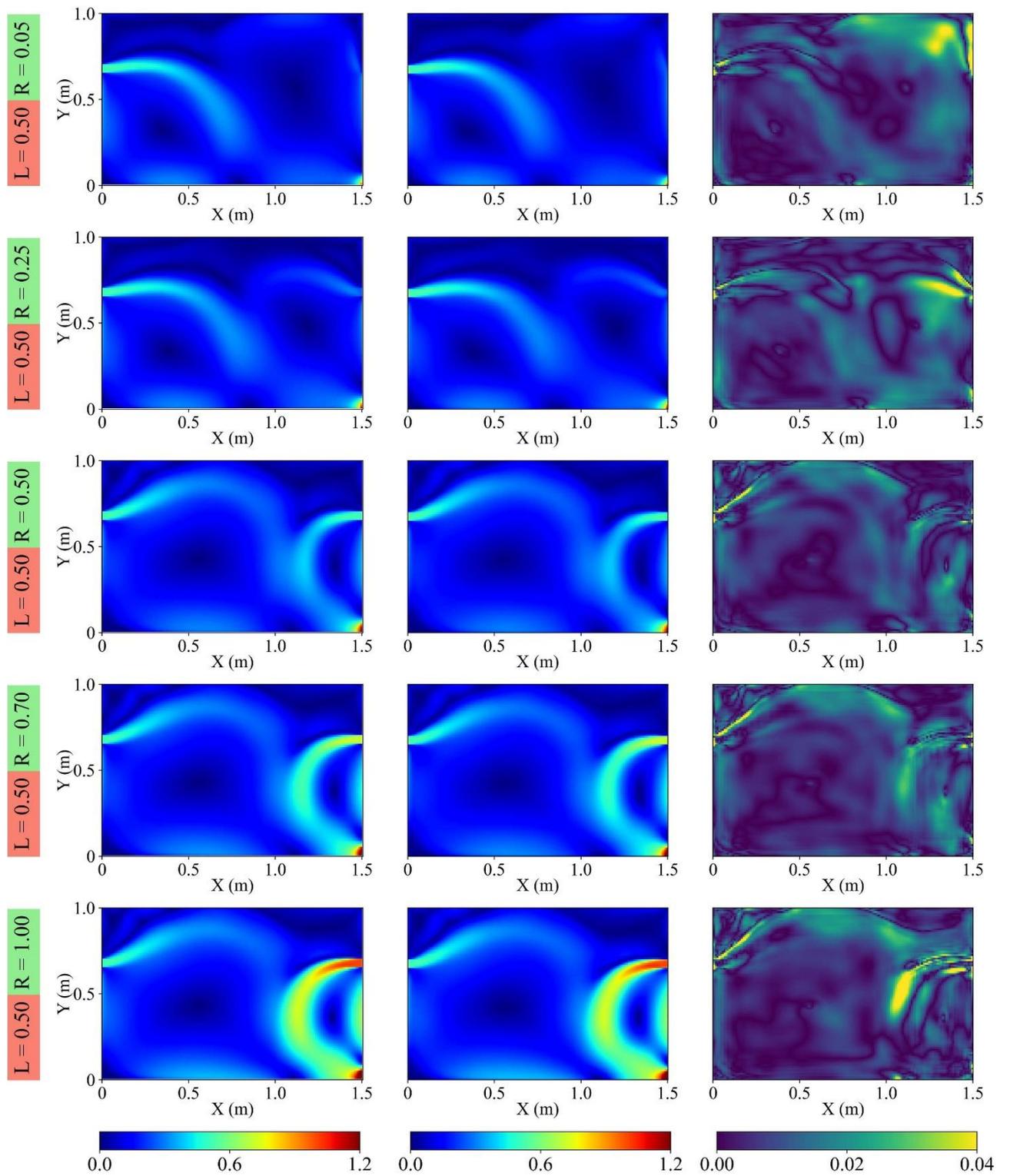

Figure 5. Training dataset of velocity fields. **Left:** Ground truth fields generated by physical CFD based on RANS; **Centre:** Predicted fields by CBML; **Right:** Absolute error.

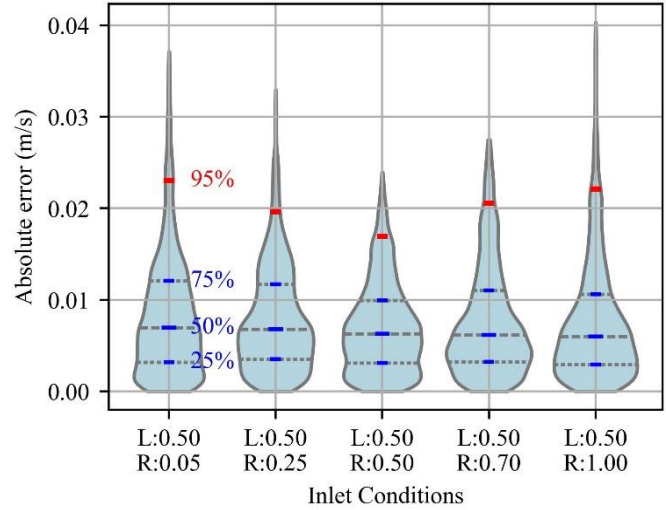

Figure 6. Violin plots of velocity absolute errors on the training dataset.

### 3.1.2 Temperature Field

In this part, reconstruction of temperature field for the same five representative on velocity training dataset is evaluated by both quality and quantity in the Figure 7 and Figure 8. As previous section, the left inlet velocity is fixed at 0.50 m/s, and the right inlet velocity varies from 0.05 m/s to 1.0 m/s. Compared to the velocity field, the temperature field is more uniform and smoother, as temperature diffusion is not only related to momentum and velocity flow, but also thermal boundary conditions and heat convection. Besides, unlike momentum transport, thermal energy transport exhibits greater inertia and diffusivity, resulting in more gradual spatial variations. Figure 7 shows the CFD ground truth, surrogate model results and spatial error map. Firstly, from a macroscopic perspective, with the increase of right inlet velocity from 0.05 m/s to 1.0 m/s, the overall indoor temperature decreases gradually, which states reasonable and accurate trend of cooling condition for a room in the summer. From perspective of details, the predicted temperature fields closely match the CFD results, with smooth transitions and well-represented thermal patterns. Across all left and right inlet combinations, the absolute error distributions are relatively uniform and of low magnitude. Most errors are occurred to thin regions near ceiling and wall boundaries, where temperature gradients are sharp due to boundary heat flux and convective cooling from inlet jets.

Figure 8 shows the density curve and quantile of reconstruction absolute error for these five representatives using violin plots. The shape of the violins is consistent across the different cases, indicating the robustness of the model across a wide range of flow and thermal boundary combinations. The median absolute errors of these five cases are less than 0.07 ℃, and it decrease with increasing right inlet velocity. The biggest absolute error shown as longest tail in the violin plot shows similar behaviours as velocity field prediction, it appears in the most asymmetric configurations (e.g., L = 0.5 m/s, R = 0.05 m/s and L = 0.5 m/s, R = 1.00 m/s), as asymmetric air supply introduces complex and irregular flow structures, such as skewed jets and unbalanced recirculation zones. Overall, the majority of spatial points exhibit small errors, confirming that the surrogate model is capable of accurately preserve spatial thermal structures and capture convective-diffusive coupling effects across a range of dual-inlet boundary scenarios.

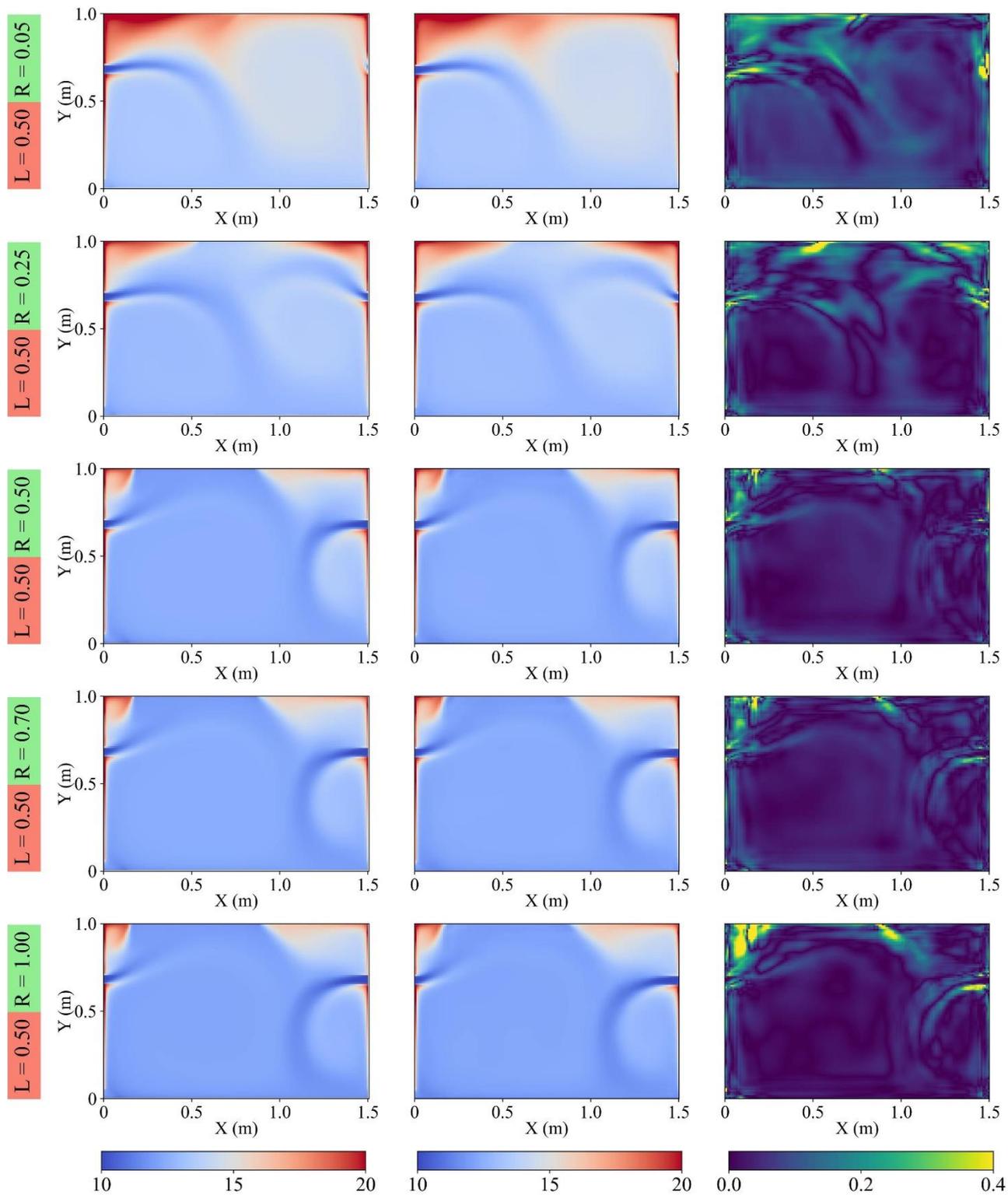

Figure 7. Training dataset of temperature fields. **Left:** Ground truth fields generated by physical CFD based on RANS; **Centre:** Predicted fields by CBML; **Right:** Absolute error.

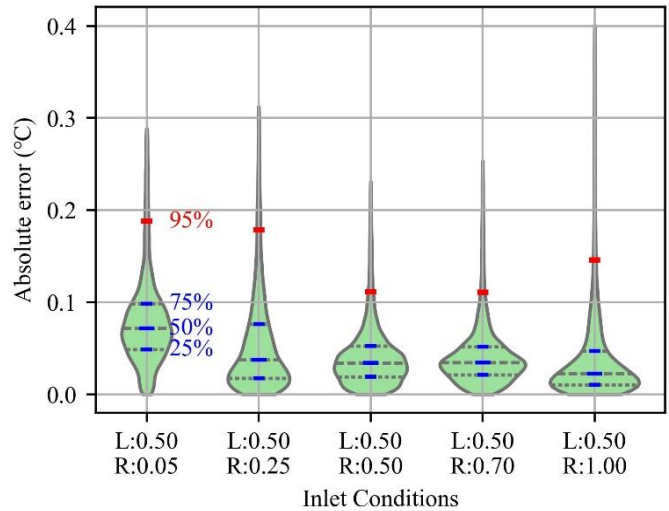

Figure 8. Violin plots of temperature absolute errors on the training dataset.

## 3.2 Prediction on testing dataset

To evaluate the generalization capability of the CBML surrogate model, a series of test cases not included in the training dataset are employed to assess its performance under unseen boundary conditions. These test cases involve new combinations of left and right inlet velocities, representing flow scenarios that were not explicitly learned during training process. In these cases, the left inlet velocity increases from 0.1 m/s to 0.9 m/s and the right inlet velocity decreases from 0.9 m/s to 0.1 m/s. Spatial velocity and temperature fields are used to provide intuitive and qualitative evaluations of the model predictions, which are further supported by quantitative analyses, including absolute error in violin plots and the correlation between predictions and CFD ground truth.

### 3.2.1 Velocity Field

The generalization performance of CBML surrogate model in predicting indoor airflow is evaluated on the testing dataset, which consists of six cases not seen during training process. Figure 9 presents the prediction results under six dual-inlet conditions. Each row shows, from left to right, the CFD reference field, the model prediction and the spatial absolute. Overall, the predicted velocity fields closely resemble the CFD results, particularly in regions with moderate velocity gradients and stable jet interactions. However, compared with the training dataset results, slightly larger deviations can be observed in regions with high velocity gradients and complex interactions between inlet streams. Besides, the spatial error map explains that largest deviations occur in cases with significant inlet velocity asymmetry (e.g., L = 0.1 m/s, R = 0.9 m/s and L = 0.9 m/s, R = 0.1 m/s), where the model struggles to fully capture the sharp momentum differences between the two inlet streams.

To further compare the error trend and statistics under different inlet conditions, violin plots are employed in Figure 10, which shows that median error of all testing cases are lower than 0.03 m/s, which is around 3.5 times higher than median error in training dataset at 0.007m/s, but the trend of errors is similar to training set, which has low errors under relative symmetry cases. Notably, the most symmetric test case shows a distribution shape comparable to that of the training set, whereas the asymmetric cases clearly deviate, supporting the observation that flow symmetry improves reconstruction stability.

Figure 11 presents scatter plots and correlation comparing CBML-predicted and CFD-computed velocity magnitudes for six representatives from the testing dataset. Each subplot corresponds to a specific dual-inlet condition, covering both symmetric and asymmetric airflow configurations. The red dashed line indicates perfect agreement, while the black dashed lines represent ±20% relative error bounds. The plots demonstrate that the surrogate model maintains strong generalization capability across test dataset, with all $R^2$ values exceeding 0.94. The cases with symmetric inlet conditions achieves the highest $R^2$ (0.9827) indicating highly accurate predictions. In contrast, the case with L = 0.1 m/s and R = 0.9 m/s, which exhibits a strong inlet imbalance, yields a slightly lower $R^2$ of 0.9410 and larger scatter around the diagonal. This performance drop is attributed to the increased complexity and asymmetry in the flow field, which poses greater challenges for the learned latent representations. Besides, the majority of predictions fall in the 20% relative error range, confirming the CBML model's robustness and its ability. However, the relative errors tend to be larger in regions with low velocity magnitudes. This is expected, as small absolute deviations in prediction can translate into higher relative errors when the true values are close to zero, which does not indicate poor model performance.

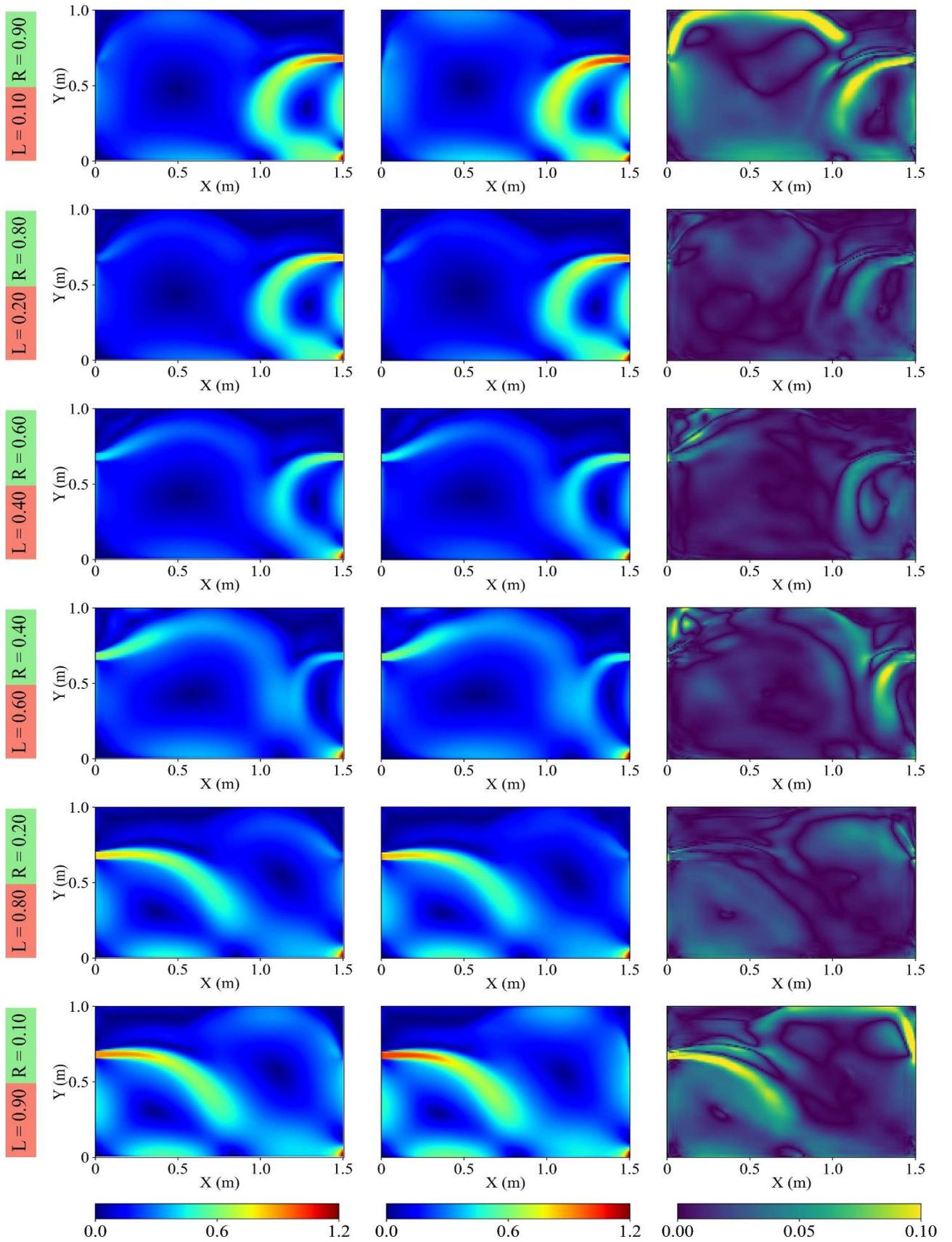

Figure 9. Test dataset of velocity fields. **Left:** Ground truth fields generated by physical CFD based on RANS; **Centre:** Predicted fields by CBML; **Right:** Absolute error.

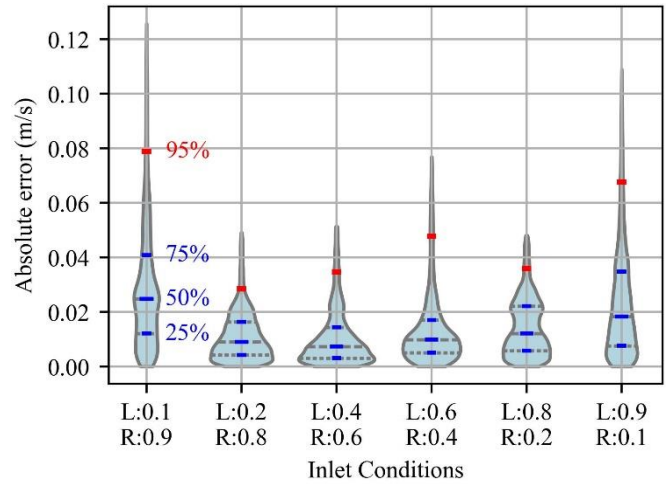

Figure 10. Violin plots of velocity absolute errors on the test dataset.

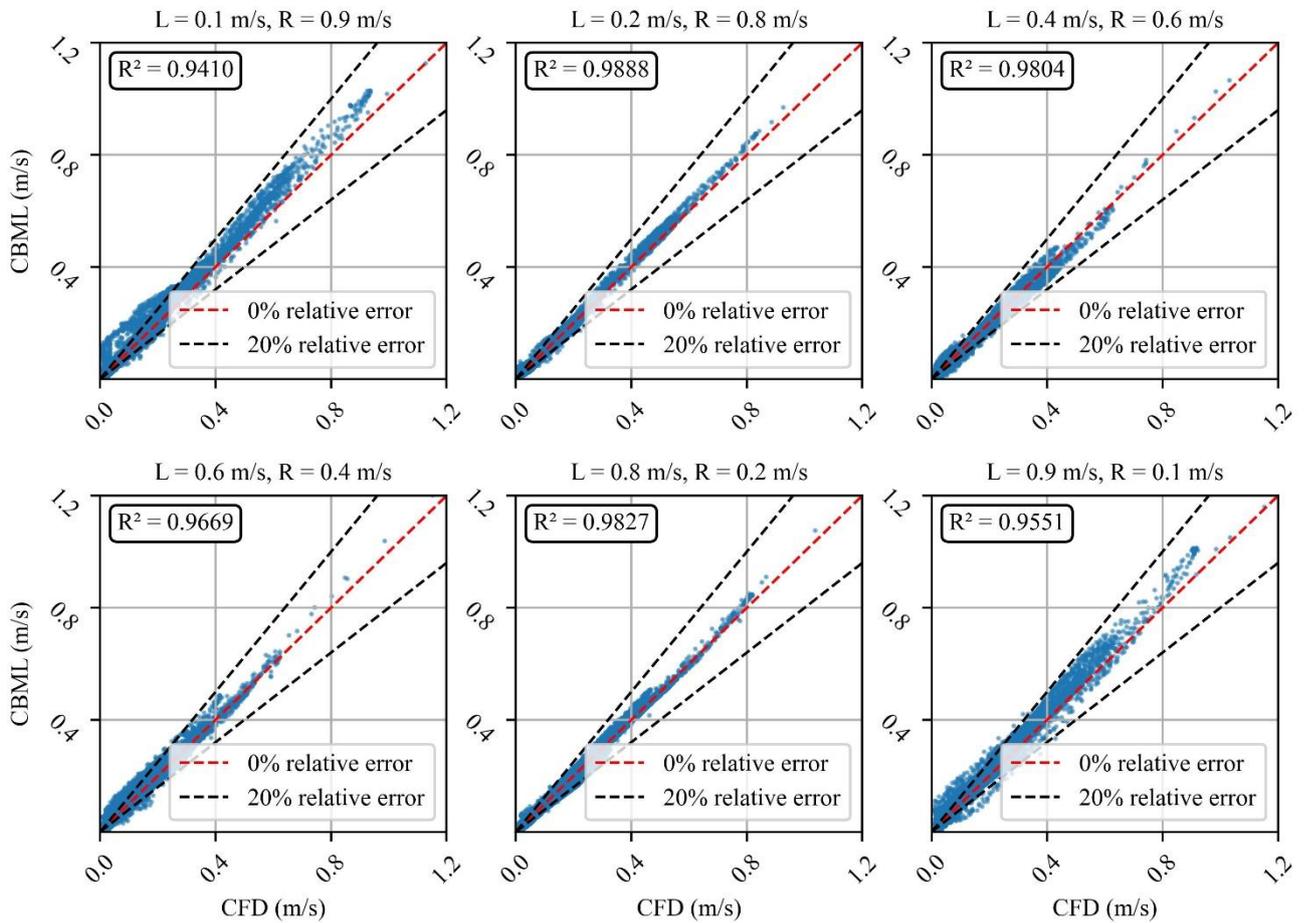

Figure 11. Scatter plots and correlation of CBML-predicted and CFD-simulated velocity field values under different inlet configurations in test dataset. The red dashed line represents perfect agreement (0% relative error), while the black dashed lines denote ±20% relative error bounds. The $R^2$ values demonstrate agreement across all cases.

### 3.2.2 Temperature Field

The surrogate model's generalization ability to predict temperature fields under unseen inlet conditions is assessed utilizing the same testing dataset as in the velocity analysis. Figure 12 shows the prediction results for six dual-inlet scenarios including CFD reference field, the predicted temperature distribution and the absolute error map. In general, the model successfully reconstructs the dominant thermal patterns and temperature distributions. The model's deviations on the training dataset (Figure 7) also occur in the test dataset: At the region in and around the boundary layers with high gradients, a drop of accuracy is observable. Besides, in the test dataset, the model has a relatively large error than training dataset in predicting the temperature of the upper area of the room, where a high temperature gradient occurs. The high gradient is caused by highest temperatures commonly observed in ceiling corner regions, where reduced ventilation and limited convective exchange result in localized accumulation of thermal energy.

Figure 13 provides a statistical summary of temperature prediction errors using violin plots. Compared to the training dataset (Figure 8), the distributions in the testing cases are generally broader and exhibit longer upper tails. The 95% error values are typically within 0.27–0.40 °C, but outliers are more frequent, especially in cases with strongly imbalanced inlet velocities. For instance, the 'L = 0.60 m/s, R = 0.40 m/s' and the 'L = 0.40 m/s, R = 0.60 m/s' cases show the longest tail and the highest 75% error, respectively, suggesting increased difficulty in capturing the heat transport induced by unstable and asymmetric velocity fields. This indicates that the model is more sensitive to localized variations in thermal boundary layers and jet-induced mixing when applied to novel flow conditions.

Figure 14 exhibits scatter and correlation between predicted temperature and CFD ground truth across the test dataset with varying inlet conditions. Overall, the surrogate model demonstrates high accuracy in temperature prediction across all test cases, with $R^2$ values exceeding 0.94. Notably, the configuration with L = 0.8 m/s and R = 0.2 m/s achieves the highest agreement ($R^2$ = 0.9960), indicating near-perfect prediction. The lowest $R^2$ of 0.9403 appears in the case of L = 0.6 m/s and R = 0.4 m/s, where the scatter becomes slightly more pronounced, particularly in low-temperature regions. Besides, this figure shows that all of the predictions fall in the range of 20% relative errors. Compared with the velocity prediction cases, the temperature prediction exhibits more compact clustering along the ideal diagonal line. This is attributed to the smoother spatial variations and the dominant role of thermal boundary conditions in determining temperature fields.

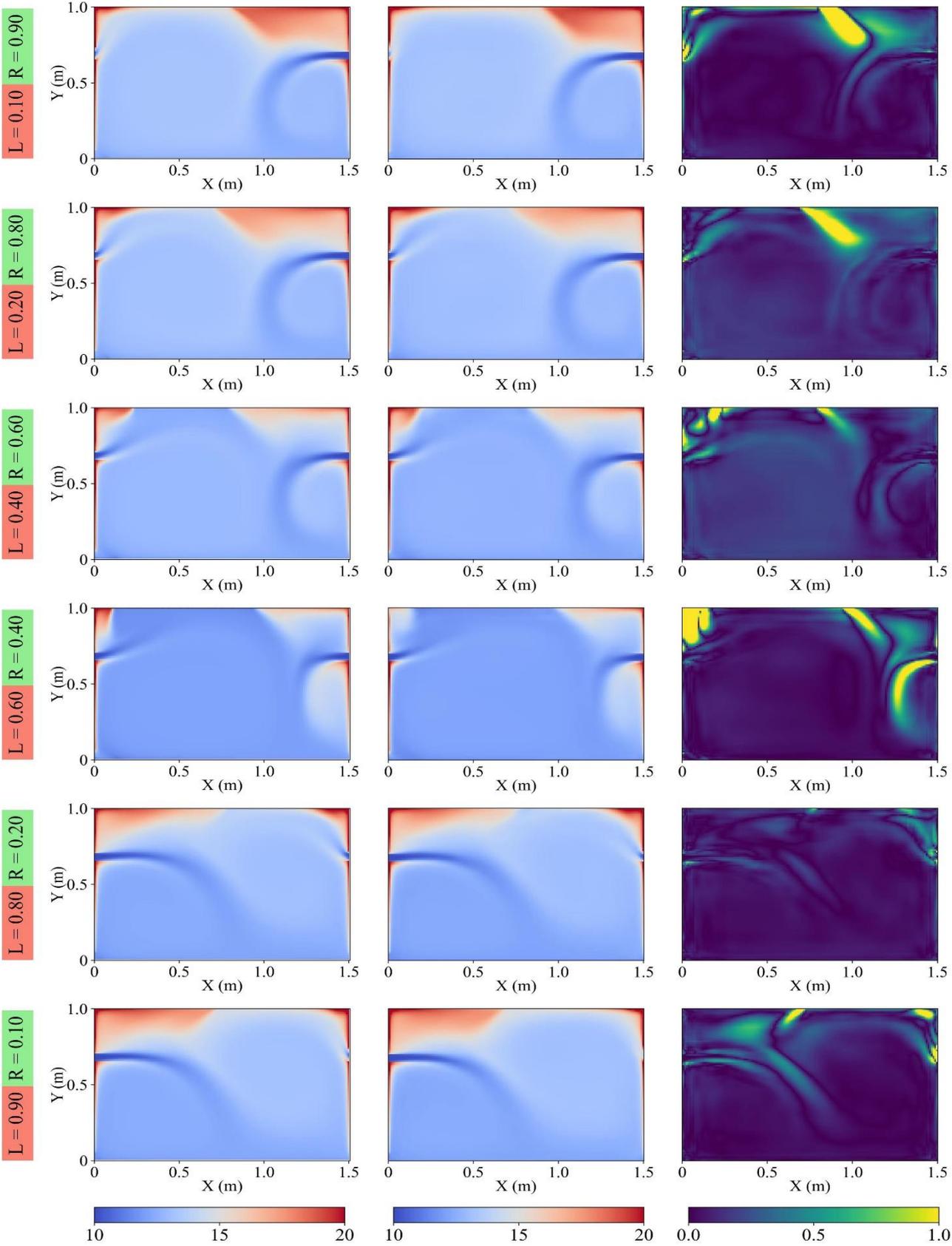

Figure 12. Test dataset of temperature fields. **Left:** Ground truth fields generated by physical CFD based on RANS; **Centre:** Predicted fields by CBML; **Right:** Absolute error.

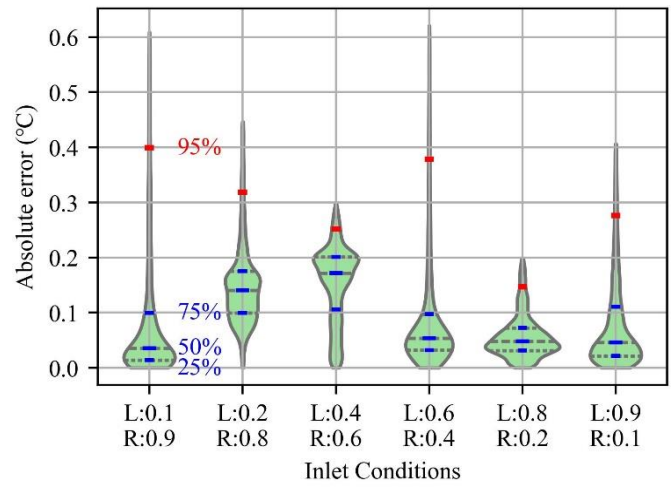

Figure 13. Violin plots of temperature absolute errors on the test dataset.

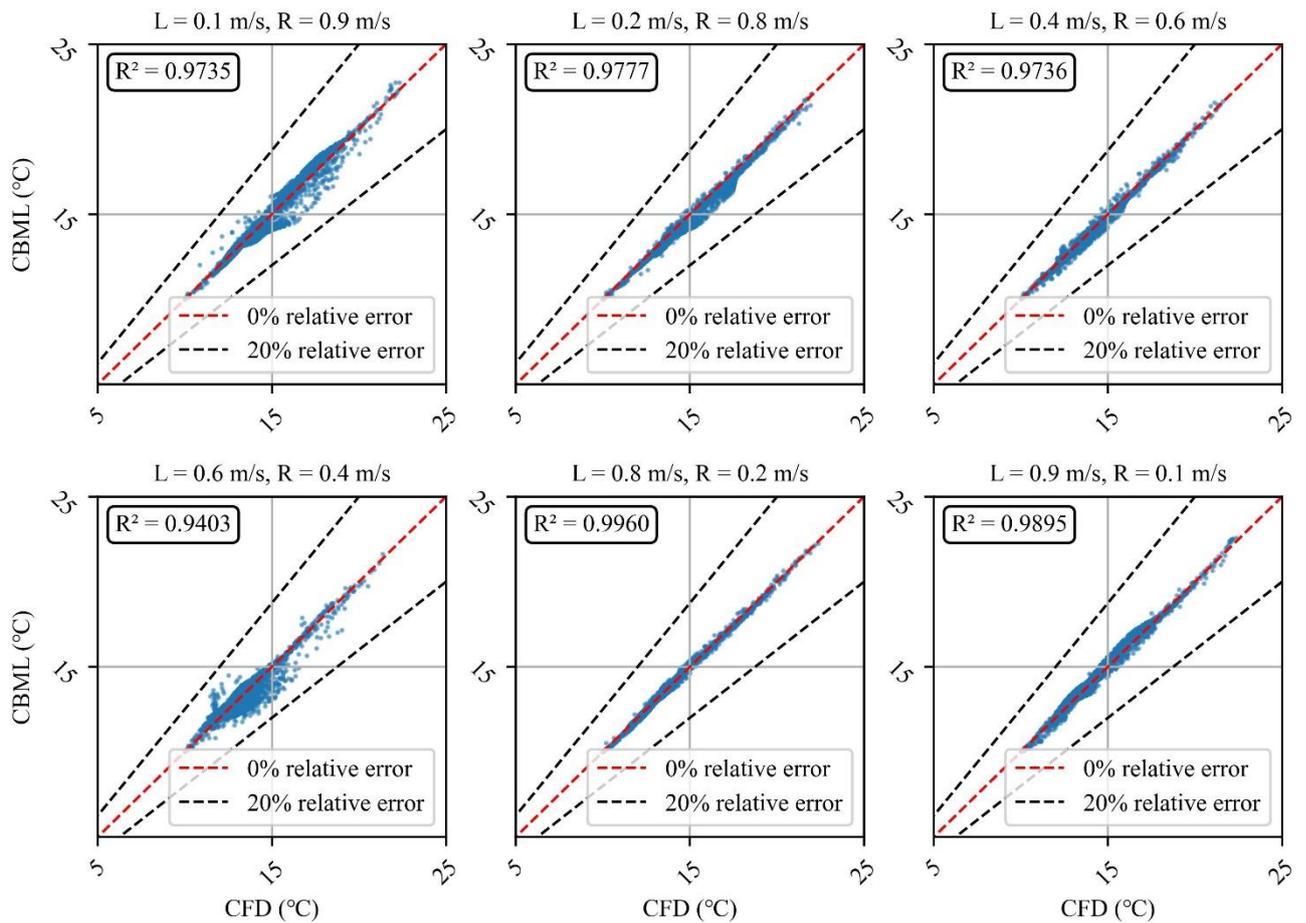

Figure 14. Scatter plots and correlation of CBML-predicted and CFD-simulated temperature field values under different inlet configurations in test dataset. The red dashed line represents perfect agreement (0% relative error),

while the black dashed lines denote ±20% relative error bounds. The R² values demonstrate agreement across all cases.

### 3.3 t-SNE Clustering Analysis

The 2D clustering visualization shown in Figure 15 illustrates how the model differentiates flow conditions based on latent features. The data points are categorized into three groups: left single-inlet (blue), right single-inlet (orange), and dual-inlet (green). From the distribution, the identified split in left-inlet and right-inlet cases as distinct clusters is evident indicating that the model effectively learns and separates their unique flow characteristics. The cluster of dual inlets is also visible as cluster and positioned between the left and right inlet. This shows that the latent space is able to capture the main flow patterns under different air supply conditions and represent their differences in a low-dimensional way. This phenomenon provides strong theoretical support for the CBML method, demonstrating that the latent space follows physical principles. It also indicates that the low-dimensional features of the dual-inlet case can be interpolated and composed from the single-inlet cases.

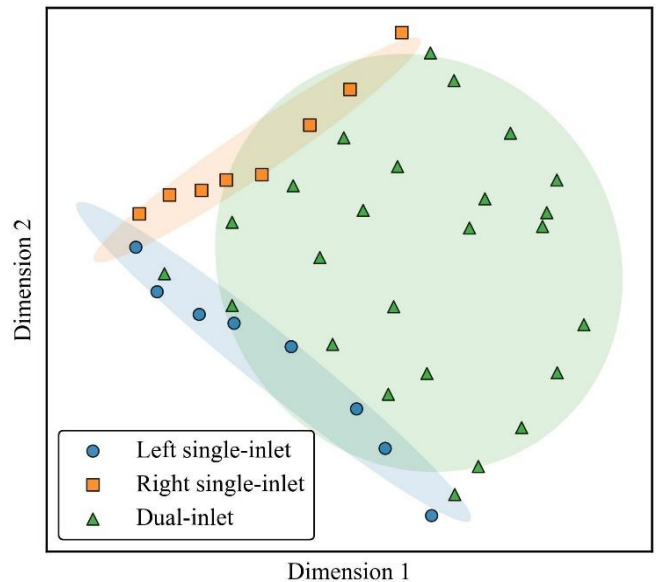

Figure 15. t-SNE Clustering for different inlet configurations in the latent space.

### 4. Conclusion

This study demonstrated that the CBML method is applicable for CFD simulation surrogate modelling to predict indoor airflow and temperature fields. The proposed surrogate model is composed of three neural network modules, a CAER as reductor, a MLP as predictor and a CNN as aggregator. A 2D room with either single or dual inlets with various velocities is employed as a benchmark case, which is simulated by an CFD model to generate the training and testing data. The results show that the model successfully learns to extract and aggregate latent features of single-component inlets (left or right) to reconstruct the combined dual-inlet flow field. The decoder of the CAER effectively maps the integrated low-rank latent representation back to full-resolution velocity and temperature fields.

Overall, the quantitative visualizations and qualitative metrics evaluations confirm the strong performance of the CBML model on both training dataset and test dataset, indicating that the neural network has effectively learned the fundamental patterns and mechanism of indoor velocity and temperature distribution. In the training set, the surrogate model accurately captures jet structures, recirculation patterns, and thermal condition, with most absolute errors falling within narrow ranges. In the test dataset, the model shows strong generalization capability, although slightly higher prediction errors are observed under strongly asymmetric inlet conditions. These deviations are attributed to increased flow complexity, amplified sensitivity to inlet imbalance, and potential representation uncertainty in the latent space. Furthermore, t-SNE clustering of the latent features reveals that the CBML model can effectively distinguish between different inlet configurations and that dual-inlet features can be meaningfully aggregated from single-inlet cases while preserving physical consistency. This supports the interpretability and modularity of the CBML approach.

In summary, the results suggest that the CBML method developed in this paper holds promise for understanding and capturing the interaction between multiple flow components based on latent space aggregation, which benefits the real-time (about on the order of 0.1 seconds) flow prediction for design and optimization of building and air conditioning systems. This approach has therefore the potential to overcome the current barrier of design-integrated CFD prediction that physical simulation and monolithic data-driven models have. A key mechanism for this purpose is the component-based approach demonstrated by the results. It supports iterative model development forming the core of a dynamic design process; it enables designers and engineers to add, remove, and modify HVAC components on the fly while getting results in real-time.

Further research will explore potential improvements, such as incorporating gradient consistency constraints to enforce better velocity transitions or expanding the training dataset to include more cases with a wider velocity range and with more HVAC components. Additionally, implementing physics-informed loss functions, such as momentum and energy constraints, could help the model better capture turbulent structures. Future work will focus on refining the surrogate model's ability to resolve sharp velocity transitions and improving its robustness across a broader range of flow conditions, ensuring its applicability to complex indoor airflow predictions.

## 5. Acknowledgement

The author gratefully acknowledges support from the Europa-Programme (Grant No. 2014/C 198/01) and the DFG Heisenberg Programme (Grant No. GE 1652/4-1).

## 6. References

[1]     Y. Lin, J. Wang, W. Yang, L. Tian, C. Candido, A systematic review on COVID-19 related research in HVAC system and indoor environment, Energy and Built Environment 5 (2024) 970–983. https://doi.org/10.1016/j.enbenv.2023.07.009.

[2]     A.N. Nair, P. Anand, A. George, N. Mondal, A review of strategies and their effectiveness in reducing indoor airborne transmission and improving indoor air quality, Environ Res 213 (2022) 113579. https://doi.org/https://doi.org/10.1016/j.envres.2022.113579.

[3]     A.M. Elsaid, M.S. Ahmed, Indoor Air Quality Strategies for Air-Conditioning and Ventilation Systems with the Spread of the Global Coronavirus (COVID-19) Epidemic: Improvements and Recommendations, Environ Res 199 (2021) 111314. https://doi.org/https://doi.org/10.1016/j.envres.2021.111314.

[4]     G.A. Ganesh, S.L. Sinha, T.N. Verma, S.K. Dewangan, Investigation of indoor environment quality and factors affecting human comfort: A critical review, Build Environ 204 (2021). https://doi.org/10.1016/j.buildenv.2021.108146.

[5]     G. Cao, H. Awbi, R. Yao, Y. Fan, K. Sirén, R. Kosonen, J. (Jensen) Zhang, A review of the performance of different ventilation and airflow distribution systems in buildings, Build Environ 73 (2014) 171–186. https://doi.org/10.1016/j.buildenv.2013.12.009.

[6]     Energyplus, (n.d.). https://energyplus.net/ (accessed May 29, 2025).

[7]     J. Clarke, Energy Simulation in Building Design, Routledge, 2007. https://doi.org/10.4324/9780080505640.

[8]     C. Chen, W.C. Lup, C. and Gorlé, Characterizing spatial variability in the temperature field to support thermal model validation in a naturally ventilated building, J Build Perform Simul 16 (2023) 477–492. https://doi.org/10.1080/19401493.2023.2179115.

[9]     P. V. Nielsen, Fifty years of CFD for room air distribution, Build Environ 91 (2015) 78–90. https://doi.org/10.1016/j.buildenv.2015.02.035.

[10]    C. Teodosiu, F. Kuznik, R. Teodosiu, CFD modeling of buoyancy driven cavities with internal heat source—Application to heated rooms, Energy Build 68 (2014) 403–411. https://doi.org/https://doi.org/10.1016/j.enbuild.2013.09.041.

[11]    C. Chen, C. Gorlé, Full-scale validation of CFD simulations of buoyancy-driven ventilation in a three-story office building, Build Environ 221 (2022). https://doi.org/10.1016/j.buildenv.2022.109240.

[12]    C. Concilio, P. Aguilera Benito, C. Piña Ramírez, G. Viccione, CFD simulation study and experimental analysis of indoor air stratification in an unventilated classroom: A case study in Spain, Heliyon 10 (2024). https://doi.org/10.1016/j.heliyon.2024.e32721.

[13]    A.A. Mansor, S. Abdullah, A.N. Ahmad, A.N. Ahmed, M.F.R. Zulkifli, S.M. Jusoh, M. Ismail, Indoor air quality and sick building syndrome symptoms in administrative office at public university, Dialogues in Health 4 (2024) 100178. https://doi.org/https://doi.org/10.1016/j.dialog.2024.100178.

[14]    A. Borrmann, P. Wenisch, C. van Treeck, E. Rank, Collaborative computational steering: Principles and application in HVAC layout, Integr Comput Aided Eng 13 (2006) 361–376. https://doi.org/10.3233/ICA-2006-13405.

[15]    P. Wenisch, C. van Treeck, A. Borrmann, E. Rank, O. Wenisch, Computational steering on distributed systems: Indoor comfort simulations as a case study of interactive CFD on supercomputers, International Journal of Parallel, Emergent and Distributed Systems 22 (2007) 275–291. https://doi.org/10.1080/17445760601122183.

[16]    Y. Zhou, S. Zheng, Z. Liu, T. Wen, Z. Ding, J. Yan, G. Zhang, Passive and active phase change materials integrated building energy systems with advanced machine-learning based climate-adaptive designs, intelligent operations, uncertainty-based analysis and optimisations: A state-of-the-art review, Renewable and Sustainable Energy Reviews 130 (2020) 109889. https://doi.org/https://doi.org/10.1016/j.rser.2020.109889.

[17] D. Kim, J.R. Jones, R.P. Schubert, E.J. Grant, D.P. Telionis, S.A. Ragab, The Application of CFD to Building Analysis and Design: a Combined Approach of an Immersive Case Study and Wind Tunnel Testing, 2013.

[18] H. Wang, Z. (John) and Zhai, Application of coarse-grid computational fluid dynamics on indoor environment modeling: Optimizing the trade-off between grid resolution and simulation accuracy, HVAC&R Res 18 (2012) 915–933. https://doi.org/10.1080/10789669.2012.688012.

[19] H. Wang, Z.J. Zhai, Analyzing grid independency and numerical viscosity of computational fluid dynamics for indoor environment applications, Build Environ 52 (2012) 107–118. https://doi.org/10.1016/j.buildenv.2011.12.019.

[20] W. Zuo, Q.Y. Chen, Validation of fast fluid dynamics for room airflow, IBPSA 2007 - International Building Performance Simulation Association 2007 (2007) 980–983.

[21] X. Han, W. Tian, J. VanGilder, W. Zuo, C. Faulkner, An open source fast fluid dynamics model for data center thermal management, Energy Build 230 (2021). https://doi.org/10.1016/j.enbuild.2020.110599.

[22] M.A.I. Khan, N. Delbosc, C.J. Noakes, J. Summers, Real-time flow simulation of indoor environments using lattice Boltzmann method, Build Simul 8 (2015) 405–414. https://doi.org/10.1007/s12273-015-0232-9.

[23] C. Caron, P. Lauret, A. Bastide, Machine Learning to speed up Computational Fluid Dynamics engineering simulations for built environments: A review, Build Environ 267 (2025). https://doi.org/10.1016/j.buildenv.2024.112229.

[24] Y. Long, X. Guo, T. Xiao, Research, Application and Future Prospect of Mode Decomposition in Fluid Mechanics, Symmetry (Basel) 16 (2024). https://doi.org/10.3390/sym16020155.

[25] Y. Lu, T. Wang, C. Zhao, Y. Zhu, X. Jia, L. Zhang, F. Shi, C. Jiang, An efficient design method of indoor ventilation parameters for high-speed trains using improved proper orthogonal decomposition reconstruction, Journal of Building Engineering 71 (2023) 106600. https://doi.org/https://doi.org/10.1016/j.jobe.2023.106600.

[26]  S. Wang, X. Chen, P. Geyer, Feasibility Analysis of POD and Deep-Autoencoder for Reduced Order Modelling: Indoor Environment CFD Prediction, in: Building Simulation Conference Proceedings, International Building Performance Simulation Association, 2023: pp. 647–654. https://doi.org/10.26868/25222708.2023.1227.

[27]  Q. Zhou, R. Ooka, Influence of data preprocessing on neural network performance for reproducing CFD simulations of non-isothermal indoor airflow distribution, Energy Build 230 (2021) 110525. https://doi.org/https://doi.org/10.1016/j.enbuild.2020.110525.

[28]  G. Liu, R. Li, X. Zhou, T. Sun, Y. Zhang, Reconstruction and fast prediction of 3D heat and mass transfer based on a variational autoencoder, International Communications in Heat and Mass Transfer 149 (2023) 107112. https://doi.org/https://doi.org/10.1016/j.icheatmasstransfer.2023.107112.

[29]  N. Morozova, F.X. Trias, R. Capdevila, E. Schillaci, A. Oliva, A CFD-based surrogate model for predicting flow parameters in a ventilated room using sensor readings, Energy Build 266 (2022) 112146. https://doi.org/https://doi.org/10.1016/j.enbuild.2022.112146.

[30]  Y. Liu, J. Wu, Z. Xu, Y. Shen, X. Guan, A hybrid mechanism-based and data-driven model for efficient indoor temperature distribution prediction with transfer learning, Energy Build 326 (2025) 115023. https://doi.org/https://doi.org/10.1016/j.enbuild.2024.115023.

[31]  S. Wang, P. Geyer, Flow Field Prediction with Various Air Inlet Positions Based on Convolutional Neural Network, in: Proceedings of the 31st International Workshop on Intelligent Computing in Engineering, EG-ICE 2024, 2024: pp. 146–155.https://www.scopus.com/inward/record.uri?eid=2-s2.0 85207181622&partnerID=40&md5=5c82e0fa52520ca2fa8a658533a55caa.

[32]  P. Geyer, S. Singaravel, Component-based machine learning for performance prediction in building design, Appl Energy 228 (2018) 1439–1453. https://doi.org/10.1016/j.apenergy.2018.07.011.

[33]   L.W. Chew, Buoyancy-driven natural ventilation: The role of thermal stratification and its impact on model accuracy, in: E3S Web of Conferences, EDP Sciences, 2023. https://doi.org/10.1051/e3sconf/202339602038.

[34]   K. Ito, K. Inthavong, T. Kurabuchi, T. Ueda, T. Endo, T. Omori, H. Ono, S. Kato, K. Sakai, Y. Suwa, H. Matsumoto, H. Yoshino, W. Zhang, J. Tu, CFD Benchmark Tests for Indoor Environmental Problems: Part 1 Isothermal/Non-Isothermal Flow in 2D and 3D Room Model, 2015. https://doi.org/https://doi.org/10.15377/2409-9821.2015.02.01.1.